\documentclass{amsart}
\usepackage{amsmath}
\usepackage{amssymb}
\usepackage{latexsym}
\usepackage{bm}
\usepackage{pstricks}
\usepackage{pst-plot}
\usepackage{psfrag}
\usepackage{graphicx}
\usepackage{float}
\usepackage{multirow}
\usepackage{tikz}
\usepackage{url}
\usepackage{multicol}
\usepackage{soul}
\usetikzlibrary{positioning}
\newdimen\nodeDist
\usepackage{xcolor}
\usepackage[percent]{overpic}

\theoremstyle{definition}

\numberwithin{equation}{section}

\newcommand{\x}{\textbf{x}}

\bibstyle{amsplain}

\begin{document}

\title[Identification of Anomalous ER Claims]{Applications of Machine Learning to the Identification of Anomalous ER Claims}


\author[J. Crawford and N. Petela]{Jesse Crawford$^{12}$ and Nicholas Petela$^{12}$}

%


%

\maketitle

\begin{center}
{\small $^1$Tarleton Analytics Institute, Tarleton State University, Stephenville, TX  76402}

{\small $^2$Department of Mathematics, Tarleton State University, Stephenville, TX  76402}
\end{center}

\begin{abstract}
Improper health insurance payments resulting from fraud and upcoding result in tens of billions of dollars in excess health care costs annually in the United States, motivating machine learning researchers to build anomaly detection models for health insurance claims.  This article describes two such strategies specifically for ER claims.  The first is an upcoding model based on severity code distributions, stratified by hierarchical diagnosis code clusters.  A statistically significant difference in mean upcoding anomaly scores is observed between free-standing ERs and acute care hospitals, with free-standing ERs being more anomalous.   The second model is a random forest that minimizes improper payments by optimally sorting ER claims within review queues.  Depending on the percentage of claims reviewed, the random forest saved  12\% to 40\% above a baseline approach that prioritized claims by billed amount. 
\end{abstract}

\section{Introduction}
The U.S. Government Accountability Office \cite{GAO} estimates that over \$50 billion dollars each year are spent on improper health insurance payments, with fraud possibly accounting for  \$18B to \$61B.  According to Che and Janusz \cite{Che2013}, the loss due to upcoding alone is approximately \$2.4 billion, prompting data analytics research in upcoding detection \cite{Bauder2016}.  He et.\ al \cite{He1997} have investigated applications of Multi-Layer Perceptron (MLP) algorithms to human labeled data from the Australian Health Insurance Commission. While they originally used four degrees of anomaly classification, they found that using only two presented a higher rate of agreement between the MLP and consultants.
Rosenburg et al.\ \cite{Rosenburg2000} built a hierarchical logistic Bayesian model based on the principal reason for admission that predicted whether audits on Diagnosis Related Group (DRG) codes were performed. This method relied on a high number of local models (one for each principal reason for admission) and achieved useful approximations of the underlying distribution of audits and related covariates, such as length of stay.

Yamanishi et al.\ \cite{Yamanishi2004} developed the SmartSifter algorithm, an on-line process for the detection of anomalous claims, by applying unsupervised machine learning to the underlying claims distribution. This method was inspired by research in network detection intrusion algorithms and was successfully applied to certain outlier detection problems in a dataset from the Australian Health Insurance Commission.
Luo and Gallagher \cite{Luo2010} also implemented an unsupervised algorithm in an Australian dataset using DRG codes, comparing individual hospitals under the assumptions that upcoding was likely adjacent (upcoded a single level) and unintentional. Their model uses decision trees to create homogenous subgroups and then tests the null hypothesis that each group originated from the same underlying DRG distribution. Che \cite{Che2013} investigated another on-line algorithm by using an adjustable time-window model to create semi-supervised rulesets. Other research includes Thornton \cite{Thornton2013}, who outlined a systematic model for claims review that includes upcoding but relies entirely on human reviewers, and Johnson \cite{Johnson2015}, who outlines a multi-step method accounting for both distance and relative density of resource expenditures by individual providers.

In Sections~\ref{upcodingsect} and \ref{costsection}, we present findings for two ER claims analytics, developed in partnership with Blue Cross and Blue Shield of Texas.  The first is an upcoding anomaly score based on ICD-10 diagnosis codes and HCPCS/CPT severity codes.  Hierarchical agglomerative clustering is applied to group diagnosis codes into risk categories, and the severity code for each claim is compared to the severity code distribution of claims within the same diagnosis code cluster.  A statistically significant difference in average upcoding anomaly scores $(P<10^{-74})$ is observed between acute care hospitals and free-standing ERs.

The second analytic addresses an optimization problem posed by the claims review process: reviewing all claims is prohibitively expensive, but reviewing an insufficient number causes fraudulent claims and improper payments to be overlooked.  The solution is to build a predictive model that  estimates cost avoidance for claims before they are reviewed and to prioritize review queues accordingly.  A random forest is applied to claims data at the header and line level and engineered features describing prescription drug abuse, doctor shopping, unnecessary unbundling, association analysis metrics, and excessive CT scans.  These features and reimbursement method anomalies are among the most predictive claim attributes, according to variable importance metrics.  The  model's performance is then evaluated in Section~\ref{econperfsect}, showing that its cost avoidance is 12\% to 40\% above a baseline approach that prioritizes claims by billed amount.  Section~\ref{intvissect} concludes with interpretation and visualization of the random forest model.

\section{Upcoding Detection}
\label{upcodingsect}


Below, we present an algorithm for assigning anomaly scores to ER claims for the purpose of upcoding detection.  The data under consideration consists of $n=\text{9,793}$ claims, indexed $i=1,\ldots,n$.  Each claim $i$ has an ICD-10 diagnosis code $d_i$ and a CPT severity code $s_i$ taking integer values from 1 (least severe) to 5 (most severe) \cite{cmsicd, cmscpt}.\footnote{These are HCPCS codes 99281 through 99285.} Upcoding is the practice of systematically assigning high severity codes to medical diagnoses where they are unwarranted \cite{Bauder2017}.  This suggests an upcoding detection strategy where each claim is compared to other claims with the same diagnosis code.  If the claim being evaluated has an unusually high severity code compared to the others, it is flagged as anomalous.

\subsection{Upcoding Anomaly Scores}
To formalize this approach, let $(D,S)$ be a randomly selected pair of diagnosis/severity codes from the entire data set $\{(d_i,s_i) \mid i=1,\ldots,n\}$.  Given a claim $(d_i,s_i)$, define the \emph{background data} to be the set of all other claims $\{(d_j,s_j) \mid j \neq i\}$.  Probability and expectation operators applied to the entire data set are denoted by $P$ and $E$ respectively, while the same operators applied to the background data are denoted $P_{(i)}$ and $E_{(i)}$.  The \emph{upcoding anomaly score} (UAS) for the $i$th claim is given by the conditional probability

\[
\text{UAS}_i = P_{(i)}(S \ge s_i \mid D = d_i).
\]

\medskip

\noindent  In other words, the upcoding anomaly score for claim $i$ is computed by first identifying all claims in the background data with the same diagnosis code $d_i$ and then computing the proportion of those claims whose severity code is greater than or equal to $s_i$.  If $s_i$ is an unusually high severity code for diagnosis $d_i$, this will be reflected by a low value of UAS$_i$ (low values are more anomalous).   The problem with this approach is high granularity of the diagnosis codes\footnote{There are 1,951 diagnosis codes present in the data.}, potentially resulting in high variance estimates of the anomaly scores.

\subsection{Hierarchical Clustering}  \label{upcodinghier}  This problem is solved by optimally reducing granularity of the diagnosis codes with a clustering algorithm.  Hierarchical clustering is well suited to dimensionality reduction of categorical variables due to its flexibility \cite{Sulc2015}.  While other clustering algorithms, such as $k$-means \cite{Forgy1965} and DBSCAN \cite{Ester96adensity-based}, make strict assumptions about the geometry or density of data clusters, hierarchical clustering can be applied to any data set where a dissimilarity metric is defined.

\noindent  To define a dissimilarity metric on the set of diagnosis codes, let

\[
\mu_{s|d} = E(S \mid D=d)
\]

\medskip

\noindent be the average severity code for all claims with diagnosis code $d$.  The dissimilarity metric between two diagnosis codes $d_1$ and $d_2$ is the distance between their corresponding average severity codes,

\[
\Delta(d_1,d_2) = |\mu_{s|d_1}-\mu_{s|d_2}|.
\]

\medskip

\begin{figure}
\begin{overpic}[width=\textwidth]{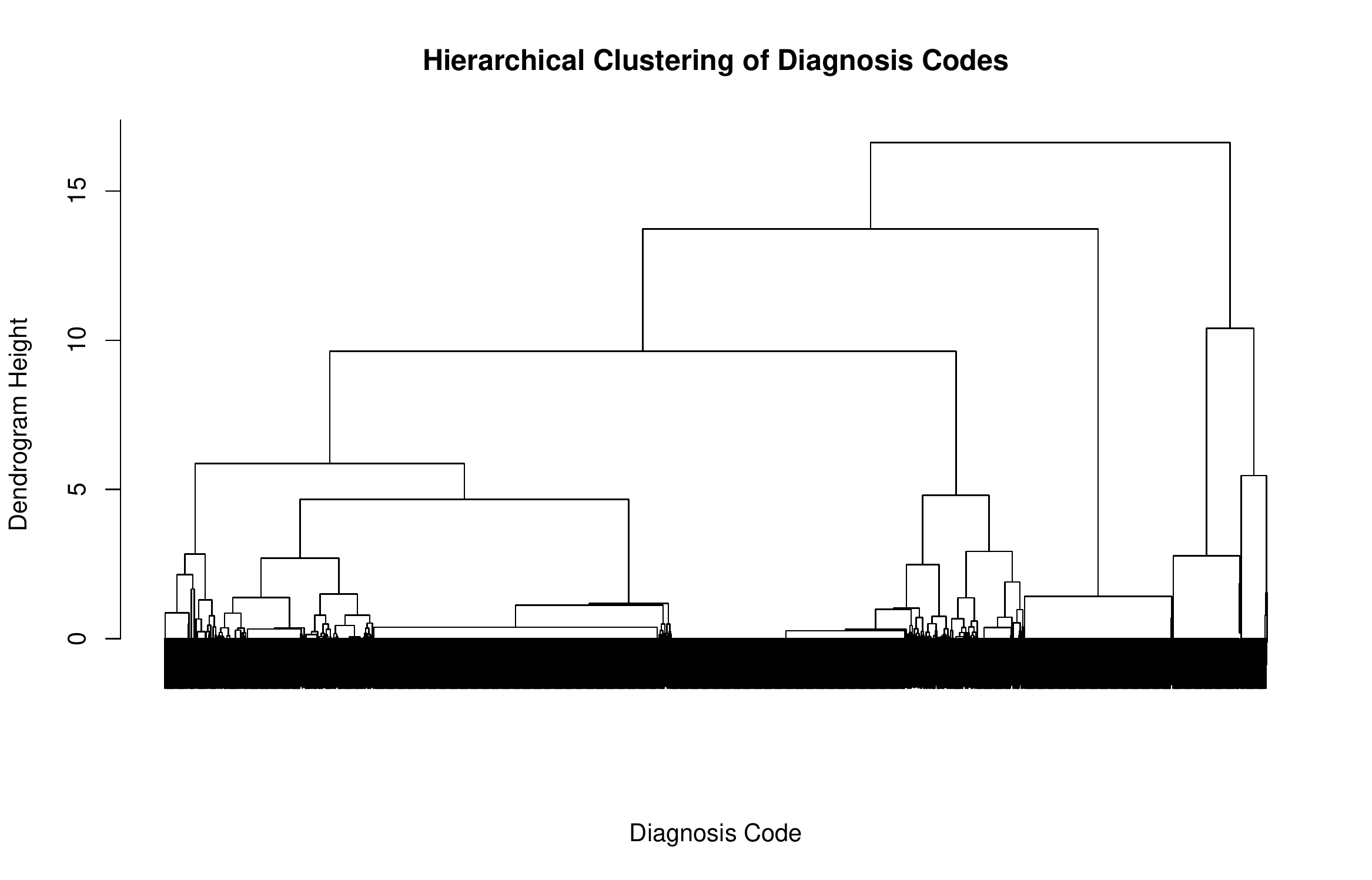}
\end{overpic}
\caption{\label{Dendrogram}Dendrogram of hierarchical clustering applied to diagnosis codes.}
\end{figure}

\noindent  Applying hierarchical clustering to this dissimilarity metric results in the dendrogram in Figure~\ref{Dendrogram}.  Each of the 1,951 diagnosis codes are represented by points at the bottom of the dendrogram.  Moving upwards, diagnosis codes that are close to each other with respect to the metric $\Delta$ are joined together into clusters, and these clusters are recursively joined to form even larger clusters, resulting in a hierarchy.

\begin{figure}
\begin{overpic}[width=\textwidth]{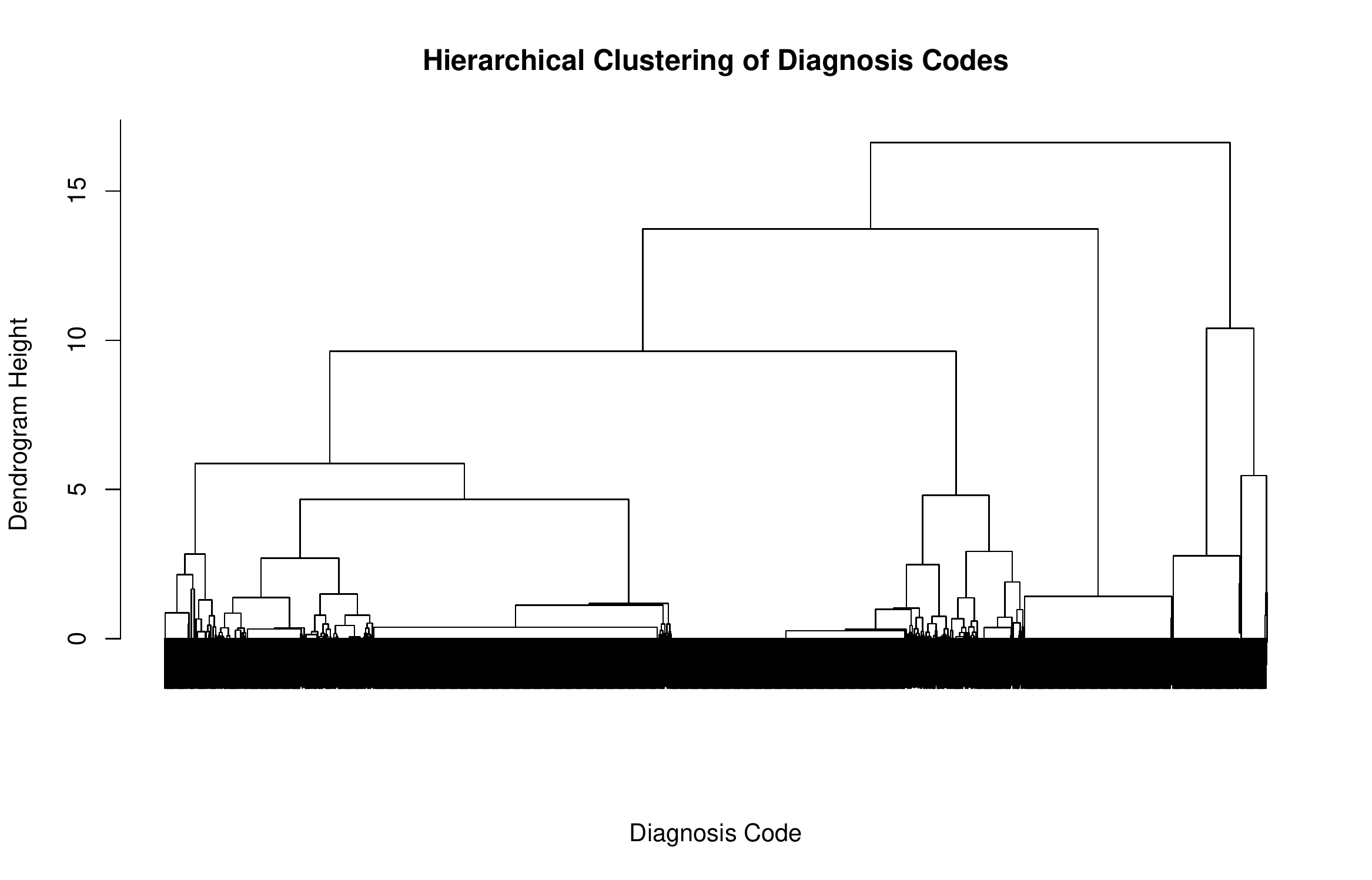}
\put(10,50){\color{blue}\rule{.83\textwidth}{1pt}}
\put(10,34.79){\color{violet}\rule{.83\textwidth}{1pt}}
\put(10,17.4){\color{red}\rule{.83\textwidth}{1pt}}
\end{overpic}
\caption{\label{Dendrogram2}Relationship between cut height and number of clusters.  Cutting at height 15 produces two clusters, cutting at height 8 produces five clusters, and cutting at height 0 produces 1,951 clusters, with each diagnosis code in a separate cluster.}
\end{figure}

 Figure~\ref{Dendrogram2} illustrates how cutting the dendrogram at various heights determines the resulting number of clusters.  For instance, cutting at height 15 would place diagnosis codes into only two clusters, causing underfitting and poor model performance \cite{Tan2018}.  Cutting at height zero places each diagnosis code in a separate cluster, causing overfitting, high variance estimates, and poor model performance \cite{Free2009}.  Instead of choosing one of these extremes, we would like to select the optimal number of diagnosis code clusters $k$ by maximizing the model performance metric described below.

First, randomly divide the claims into two subsets $A$ and $B$ for the purpose of two-fold cross-validation.\footnote{Initially, $A$ will serve as a training set to build a model and $B$ will serve as a test set to compute model performance, and then the roles of $A$ and $B$ will be reversed.  After both steps are complete, the model performance scores are averaged.}   Consider a dendrogram cut resulting in $k$ clusters, let $d_i^{(k)}$ denote the $i$th claim's diagnosis code cluster, and let $(D^{(k)},S)$ be randomly selected from the training data. If claim $i$ is in the test data, then the  probability that its severity code is equal to $s$ is estimated by

\begin{enumerate}
\item
identifying all claims in the training data with the same diagnosis code cluster $d_i^{(k)}$ and 
\item
computing the proportion of those claims with severity code $s$, that is,
\end{enumerate}

\[
\hat{p}_i(s)= P\left(S=s \mid D^{(k)} =  d_i^{(k)}\right)\text{, for }s=1,\ldots,5.
\]

\medskip

  \noindent In this manner, each claim $i$ in the test data is assigned a vector of predicted probabilities \\ $(\hat{p}_i(1),\ldots,\hat{p}_i(5))$, and these vectors are compared to the true severity codes $s_i$ to compute model performance (ordinal AUC) \cite{OrdAUC}.  Repeating this process for both folds $A$ and $B$ and averaging the results yields the ordinal AUC for a dendrogram cut resulting in $k$ clusters.  

\begin{figure}
\begin{center}
\includegraphics[width=9.5cm]{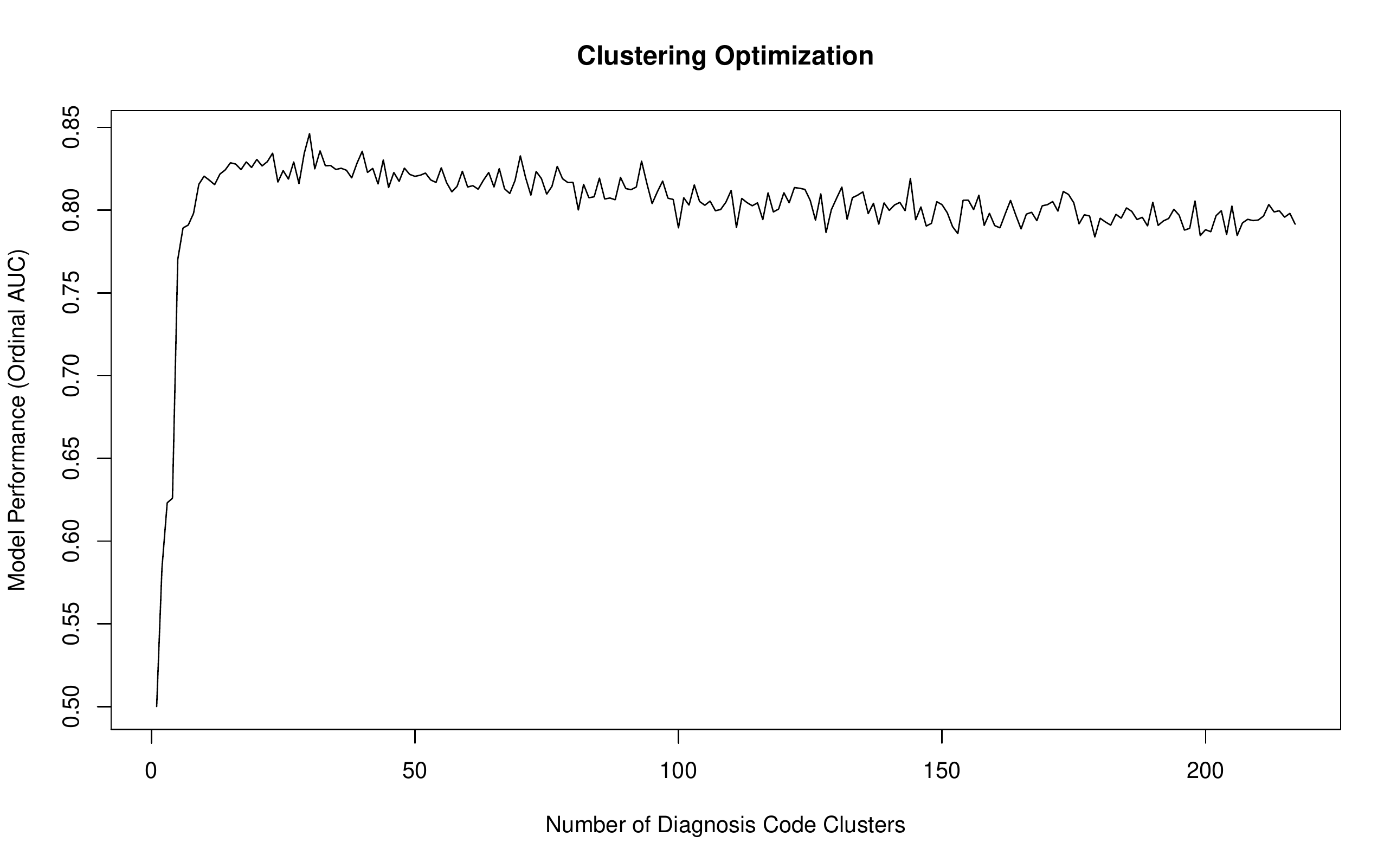}
\end{center}

\vspace{-7mm}

\caption{\label{clustopt} Ordinal AUC vs Number of Diagnosis Code Clusters $k$.}
\end{figure}

\begin{figure}
\begin{center}
\includegraphics[width=9.5cm]{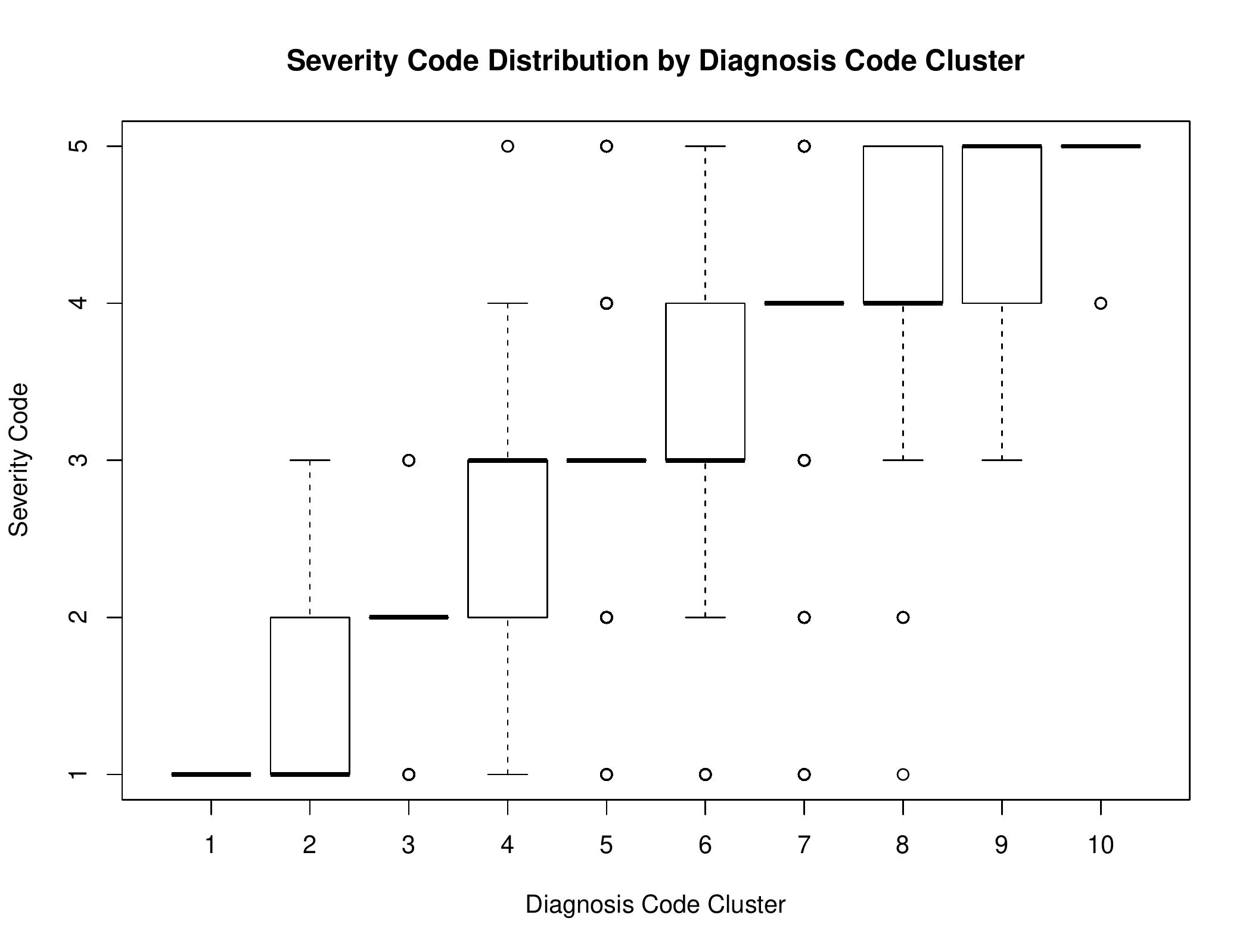}
\end{center}

\vspace{-7mm}

\caption{\label{sevclust} Box plot of severity code stratified by diagnosis code cluster.}
\end{figure}

Figure~\ref{clustopt} shows that the maximum ordinal AUC (0.846) occurs at $k=30$, however, this yields some clusters with samples sizes of six claims, which risks unstable model estimates.  Decreasing the number of clusters to $k=10$ achieves sample sizes greater than 30 for all clusters while only decreasing ordinal AUC to 0.821.  Figure~\ref{sevclust} displays box plots of severity codes stratified by diagnosis code cluster, showing that this clustering effectively partitions diagnosis codes into different severity code profiles.

\subsection{Upcoding Anomaly Scores Revisited}

After optimally reducing granularity of the diagnosis codes, we are prepared to improve the  upcoding anomaly scores.  Previously, the anomaly score for claim $i$ was based on claims in the background data with the same diagnosis code $d_i$.  The key modification is to consider claims in the background data with the same \emph{diagnosis code cluster} $d^{(k)}_i$:

\begin{equation}
\label{mUAS}
\text{UAS}_i = P_{(i)}(S \ge s_i \mid D^{(k)} = d^{(k)}_i).
\end{equation}

\medskip

\noindent  An additional modification is useful for stratifying the anomaly scores based on the levels of a categorical predictor $X$.  Suppose $X$ has levels $m=1,\ldots,M$, and let $m(i)$ be the level of $X$ for the $i$th claim.  Then the background data for claim $i$ is all claims $j$ such that $m(i)\neq m(j)$.  For instance, to compare upcoding anomaly scores for acute care hospitals and  free-standing ERs, the background data would be defined as follows:

\begin{itemize}
\item
if claim $i$ is from an acute care hospital, its background data consists of all claims from free-standing ERs
\item
if claim $i$ is from a free-standing ER, its background data consists of all claims from acute care hospitals.
\end{itemize}

\begin{figure}
\begin{center}
\includegraphics[width=\textwidth]{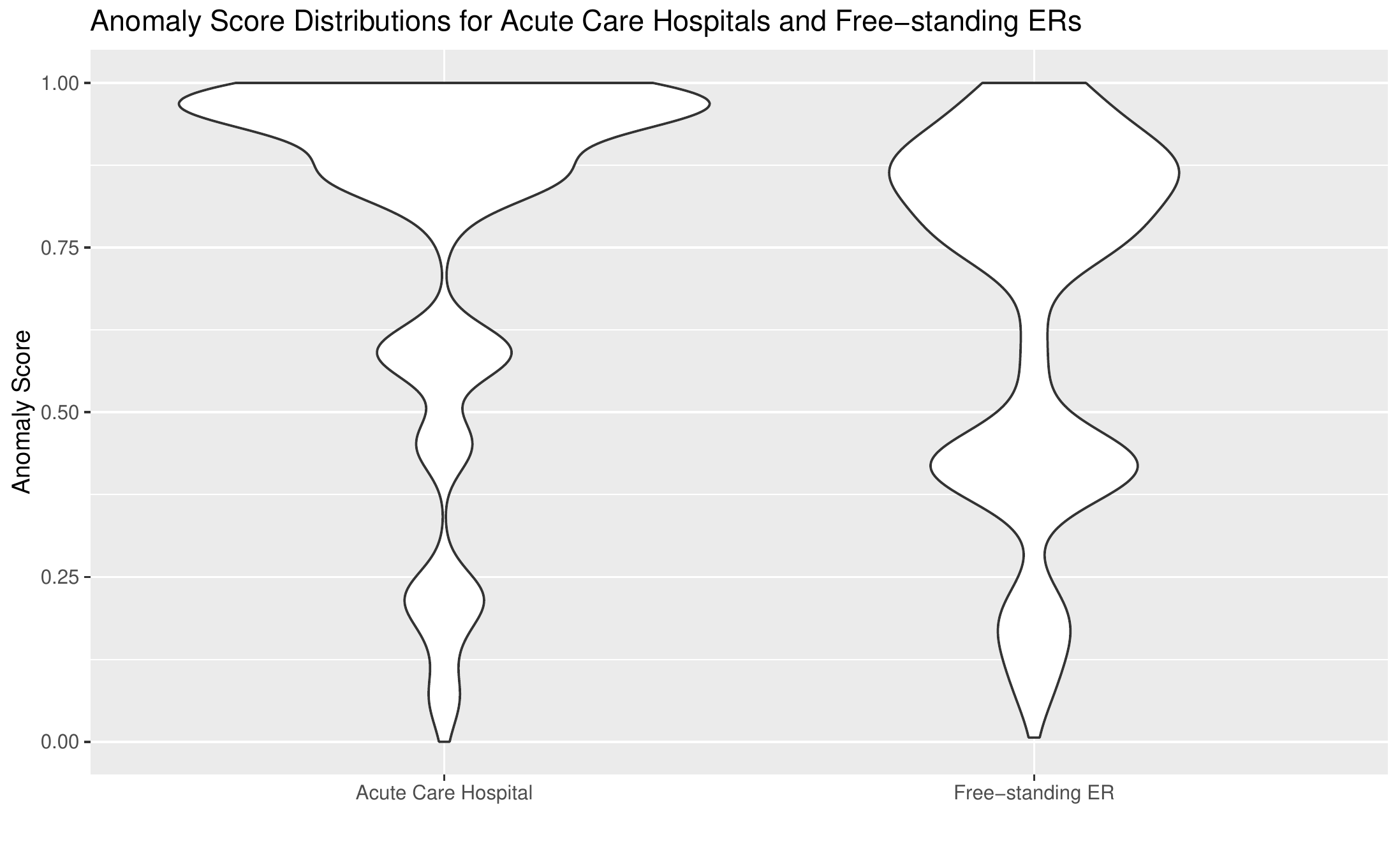}
\end{center}

\caption{\label{fseranomaly} Upcoding anomaly score distributions for acute care hospitals and free-standing ERs.}
\end{figure}

\noindent  With this modification in place, Equation~\eqref{mUAS} is used to calculate upcoding anomaly scores, with the understanding that $P_{(i)}$ is applied to the modified background data.  Keeping in mind that smaller values of UAS are more anomalous, Figure~\ref{fseranomaly} shows a greater concentration of moderately anomalous claims (in the 0.25 to 0.5 range) at free-standing ERs.  In fact, the mean UAS at acute care hospitals and free-standing ERs are 0.79 and 0.66, respectively, and this difference is highly statistically significant ($P< 10^{-74}$).

\section{Cost Avoidance Optimization}
\label{costsection}

\noindent  The previous section described a method for detecting one specific type of health insurance anomaly, upcoding, on the basis of two predictor variables, diagnosis and severity codes.  This section presents a more extensive model for detecting several types of health insurance anomalies within the unified framework of cost avoidance optimization.  The data consist of $n=\text{20,086}$ ER claims that have been inspected by claims reviewers.  The cost avoidance for a claim is the difference between its claim amounts before and after review:

\[
\text{Cost Avoidance} = (\text{Prereview Claim Amount}) - (\text{Postreview Claim Amount})
\]

\medskip

\noindent   Our objective is to predict  cost avoidance \emph{before} a claim is reviewed using the following explanatory variables, which are defined for the claim (header level) or for individual transaction lines on a claim (line level).

\begin{itemize}
\item 
\textbf{Billed Amount}
\item 
\textbf{Number of Claim Lines}
\item 
\textbf{ICD-10 Diagnosis Code}
\item
\textbf{ICD-10 Procedure Code}
\item 
\textbf{HCPCS/CPT Code}
\item 
\textbf{HCPCS/CPT Modifier Codes}
\item
\textbf{Facility Observational Units}
\item
\textbf{Product Type}
\item
\textbf{Network Status}.  In/out of network. 
\item 
\textbf{Revenue Code}. Used on institutional claims to identify the type of service or procedure.
\item 
\textbf{Acute/Ancillary Code}, e.g., acute care hospital, or free-standing ER.
\item 
\textbf{Servicing Provider Type}, e.g., medical doctor, dentist, or optician.
\item
\textbf{Servicing Provider Specialty}, e.g., obstetrics, pediatrics,  or anesthesiology.
\item 
\textbf{Servicing Provider Key}.  Identifies each provider for the purpose of estimating fixed effects at the provider level. 
\item 
\textbf{Place of Treatment Code.}  Identifies the place of treatment where health care services were rendered.

\item
\textbf{Institutional Pricing Code}.  Indicates if the service is contained within the facility allowed amount or is subject to financial carve-outs.
\item 
\textbf{Reimbursement Method}, e.g., percent of charge or flat rate.
\end{itemize}

\noindent   The model also bases its predictions on features that have been engineered to detect the following types of anomalies:

\begin{itemize}
\item
Prescription Drug Abuse
\item
Doctor Shopping
\item
Unnecessary Unbundling
\item
Association Analysis Metrics
\item
Excessive CT Scans.  
\end{itemize}

\subsection{Hierarchical Clustering for the Cost Avoidance Model}  \label{hiercamodel} As discussed in Section~\ref{upcodinghier}, hierarchical clustering is used to reduce dimensionality of the categorical predictor variables at the header and line levels.  The main difference for the cost avoidance model is the measurement scale of the dependent variables.  Cost avoidance is quantitative, while severity codes are ordinal.  

Given a categorical variable $D$, e.g. diagnosis code, we first fit an ANOVA model for predicting cost avoidance in terms of $D$.  For any two possible values $d_1$ and $d_2$ of $D$, let $T(d_1,d_2)$ denote the $t$-statistic for testing equality of their coefficients in the ANOVA model.  The dissimilarity metric is then simply the absolute value of this $t$-statistic:

\[
\Delta(d_1,d_2) = |T(d_1,d_2)|.
\]

\medskip

\begin{figure}

\begin{center}
\includegraphics[width=\textwidth]{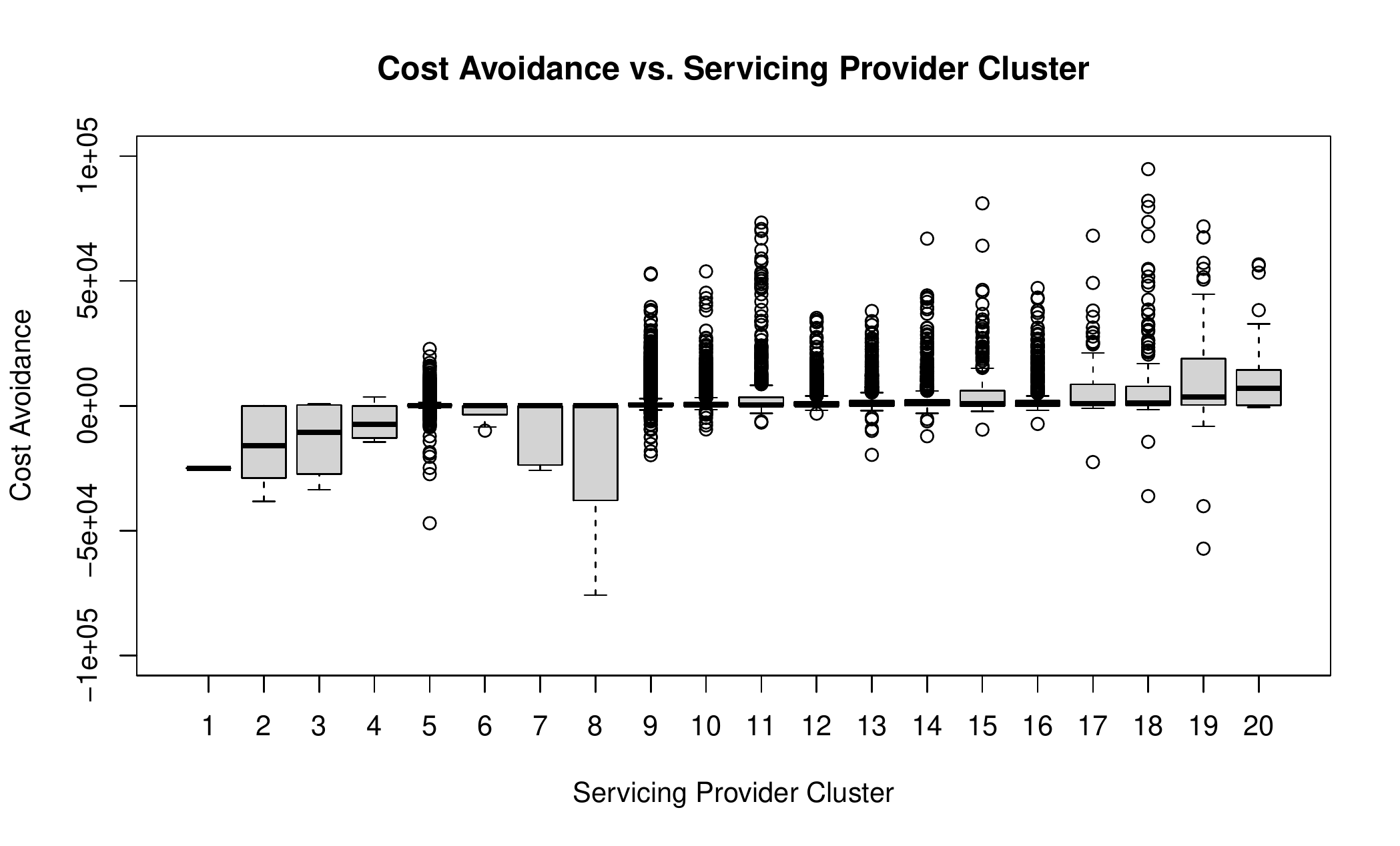}
\end{center}

\caption{\label{ProvCluster} Box plot of cost avoidance stratified by servicing provider cluster.  Servicing providers in clusters 1 through 8 tend to understate ER claims, while clusters 9 through 20 tend to overstate them.}
\end{figure}

\noindent  With this dissimilarity metric in place, hierarchical clustering proceeds as previously described.  First, a dendrogram is constructed, and then the number of clusters is determined by optimizing model performance, the coefficient of determination $R^2$, with two-fold cross-validation.   For example, the 361 servicing provider keys represented in the data are binned into only 20 servicing provider clusters, as displayed in Figure~\ref{ProvCluster}, demonstrating an important benefit of the clustering algorithm beyond dimensionality reduction.  The servicing provider clusters can be viewed as an analytic that groups providers into different risk categories.  At this point, it should be mentioned that cost avoidance can be negative in cases where a claim review resulted in an increased claim payment, and machine learning can identify these claims to ensure that patients receive the claim they are entitled to.  Figure~\ref{ProvCluster} reveals that servicing providers in clusters 1 through 8 tend to understate ER claims, while clusters 9 through 20 tend to overstate them.  


\newpage

\subsection{Building the Cost Avoidance Model with Random Forests} \label{rfsect} Once hierarchical clustering is complete, we are prepared to build the cost avoidance model.  The data is $\{(\text{CA}_i,\x_i) \mid i=1,\ldots,n\}$, where $\text{CA}_i$ is the $i$th claim's cost avoidance, and $\x_i = (x_{i1},\ldots,x_{ip})$ is the vector of header/line data and engineered features for the $i$th claim.  The goal is to build a model for predicting $\text{CA}_i$ in terms of $\x_i$, and this is achieved in three steps:  cost avoidance downscaling, random forest predictions, and cost avoidance upscaling.

\begin{enumerate}
\item
Define the cost avoidance ratio (CAR) of the $i$th claim to be the ratio of its cost avoidance to the billed amount:

\[
\text{CAR}_i = \frac{\text{CA}_i}{(\text{Billed Amount})_i}.
\]

\medskip

\noindent  Cost avoidance is always less than or equal to billed amount, i.e., $\text{CAR}_i \le 1$, and the predictive model needs to preserve this feature.

\medskip

\item
A random forest model $\widehat{\text{CAR}}_i = \text{RF}(\x_i)$ is then built for predicting $\text{CAR}_i$ using the explanatory variables $\x_i$.

\item
The random forest predictions are then upscaled to produce cost avoidance predictions:

\[
\widehat{\text{CA}}_i = (\widehat{\text{CAR}}_i)(\text{Billed Amount})_i.
\]

\end{enumerate}
\medskip

%
%
%
%
%

\begin{figure}
\begin{tikzpicture}[
    node/.style={%
      draw,
      rectangle,
    },
  ]

    \node [node] (A) {$x_{j_1} \ge c_1$?};
    \path (A) ++(-135:\nodeDist) node [node] (B) {Leaf Node 1};
    \path (A) ++(-45:\nodeDist) node [node] (C) {$x_{j_2} \ge c_2$?};
    \path (C) ++(-135:\nodeDist) node [node] (D) {Leaf Node 2};
    \path (C) ++(-45:\nodeDist) node [node] (E) {Leaf Node 3};

    \draw (A) -- (B) node [left,pos=0.25] {Yes}(A);
    \draw (A) -- (C) node [right,pos=0.25] {No}(A);
    \draw (C) -- (D) node [left,pos=0.25] {Yes}(A);
    \draw (C) -- (E) node [right,pos=0.25] {No}(A);
\end{tikzpicture}

\caption{\label{Decisiontree} Example of a decision tree with two binary splits.}
\end{figure}
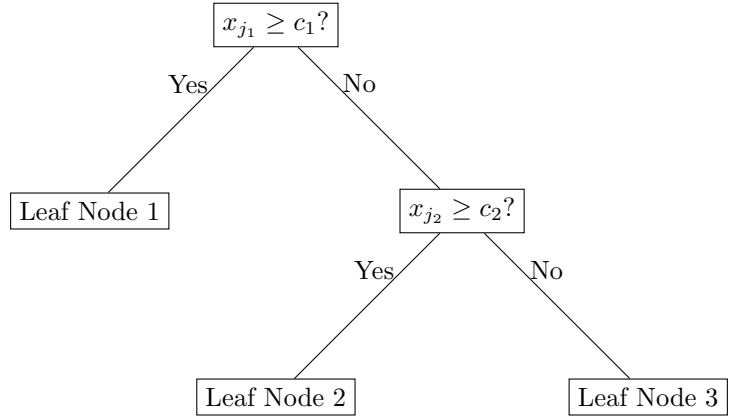

\begin{figure}
\begin{center}
\includegraphics[width=12cm]{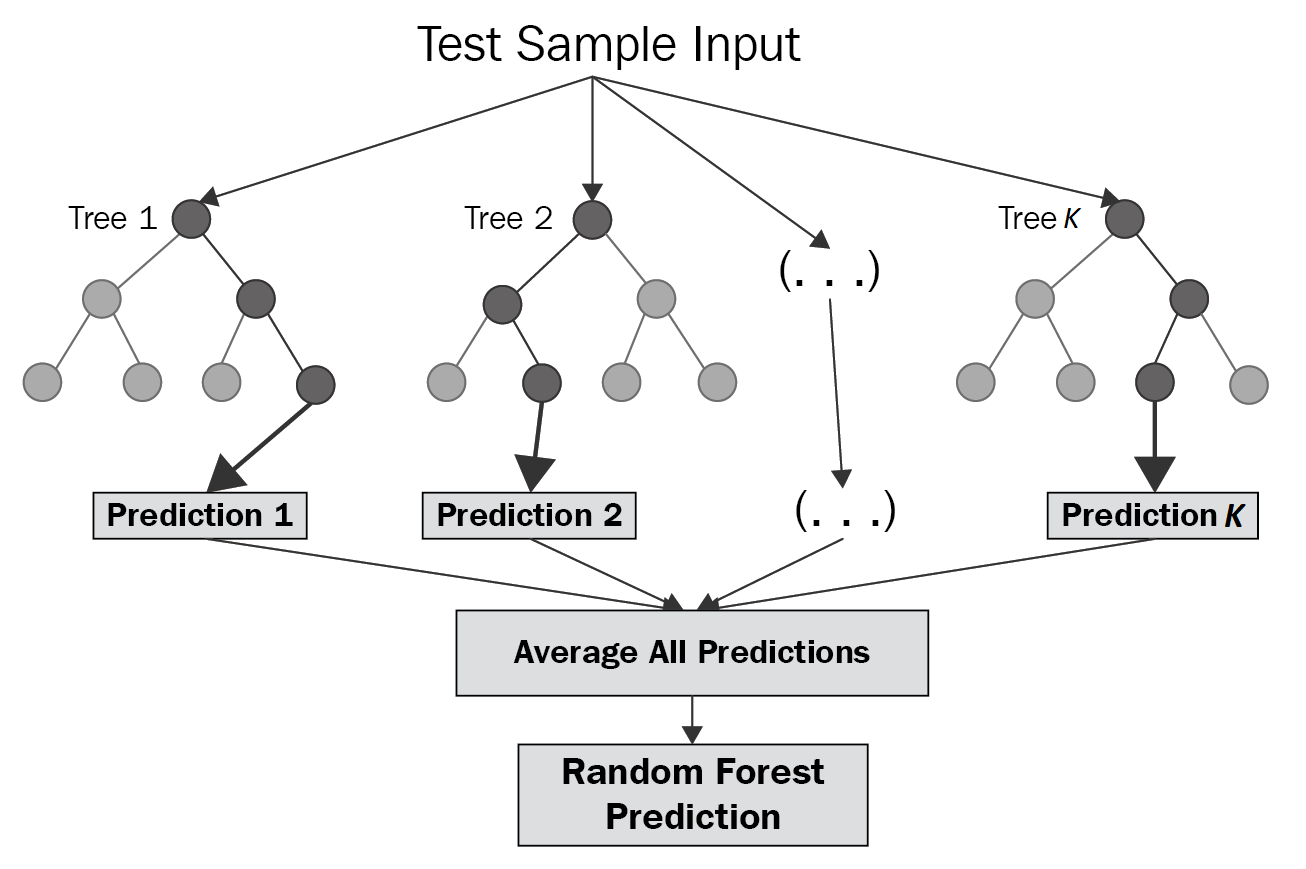}
\end{center}
\caption{\label{RFfigure} Random forests are ensemble models based on decision trees \cite{rfimg}.}
\end{figure}

\noindent  The random forest model described in step (2) is built as an ensemble of many decision trees, as illustrated in Figure~\ref{RFfigure}.  Decision trees are essentially flow charts, built by recursively partitioning the training data $(\text{CAR}_i,\x_i)$, $i=1,\ldots,n$, based on the values of the independent variables.  When the independent variables $x_j$ are quantitative, the splits can be binary, as displayed in Figure~\ref{Decisiontree}, or multiway splits can be used by binning the values of $x_j$.  Any given split partitions the data into several nodes, and the \emph{impurity} of each node is defined to be the sum of square deviations from the mean
 for the values of $\text{CAR}$ in that node.  Decision tree algorithms, such as AID and CART \cite{Morgan1963, Breiman1984}, generate splits which minimize the sum of the impurity measures for the resulting nodes until a stopping criterion is met.  When presented with a new test case $\x$, the decision tree classifies it into one of the leaf nodes, and the estimate $\widehat{\text{CAR}}=f(\x)$ is determined by fitting a linear model to that node.

Random forests \cite{Breiman2001} are constructed by first generating a sequence of independent and identically distributed random vectors $\Theta_1,\ldots,\Theta_K$, which are used to build $K$ decision trees, $f_k = f(\cdot, \Theta_k)$.  The random forest then produces estimates for the cost avoidance ratio by averaging the estimates of these decision trees,

\[
\widehat{\text{CAR}} = \text{RF}(\x) = \frac{1}{K} \sum_{k=1}^K   f(\x,\Theta_k).
\]

\noindent  The randomness from the vectors $\Theta_1,\ldots,\Theta_K$ can influence growth of the trees in several ways.  In bootstrap aggregating (bagging), the training data for each tree is a random sample with replacement from the original training data.  Other ways to introduce randomness include randomizing each tree's input features $x_j$, or randomly choosing each decision tree split from a set of top performing splits.

 As the number of trees tends to infinity, the mean square error of a random forest converges to $E(y-E_\Theta f(\x,\Theta))^2$, where $y=\text{CAR}$, and $E_{y,\x}$ and $E_\Theta$ are expectations with respect to the probability distributions of $(y,\x)$ and $\Theta$, respectively \cite{Breiman2001}.  Therefore, there is no danger of overfitting when increasing the number of trees in a random forest, but one should ensure that a sufficient number of trees are being used to achieve high performance.

\subsection{Economic Performance of the Cost Avoidance Model}  \label{econperfsect}  As discussed in the introduction, reviewing all claims is not feasible, but reviewing too few risks overlooking a large number of  fraudulent claims and improper payments.  To solve this problem, claims in review queues should be sorted by predicted cost avoidance $\widehat{\text{CA}}$, so that high priority claims are reviewed first.

\begin{figure}
\includegraphics[width=\textwidth]{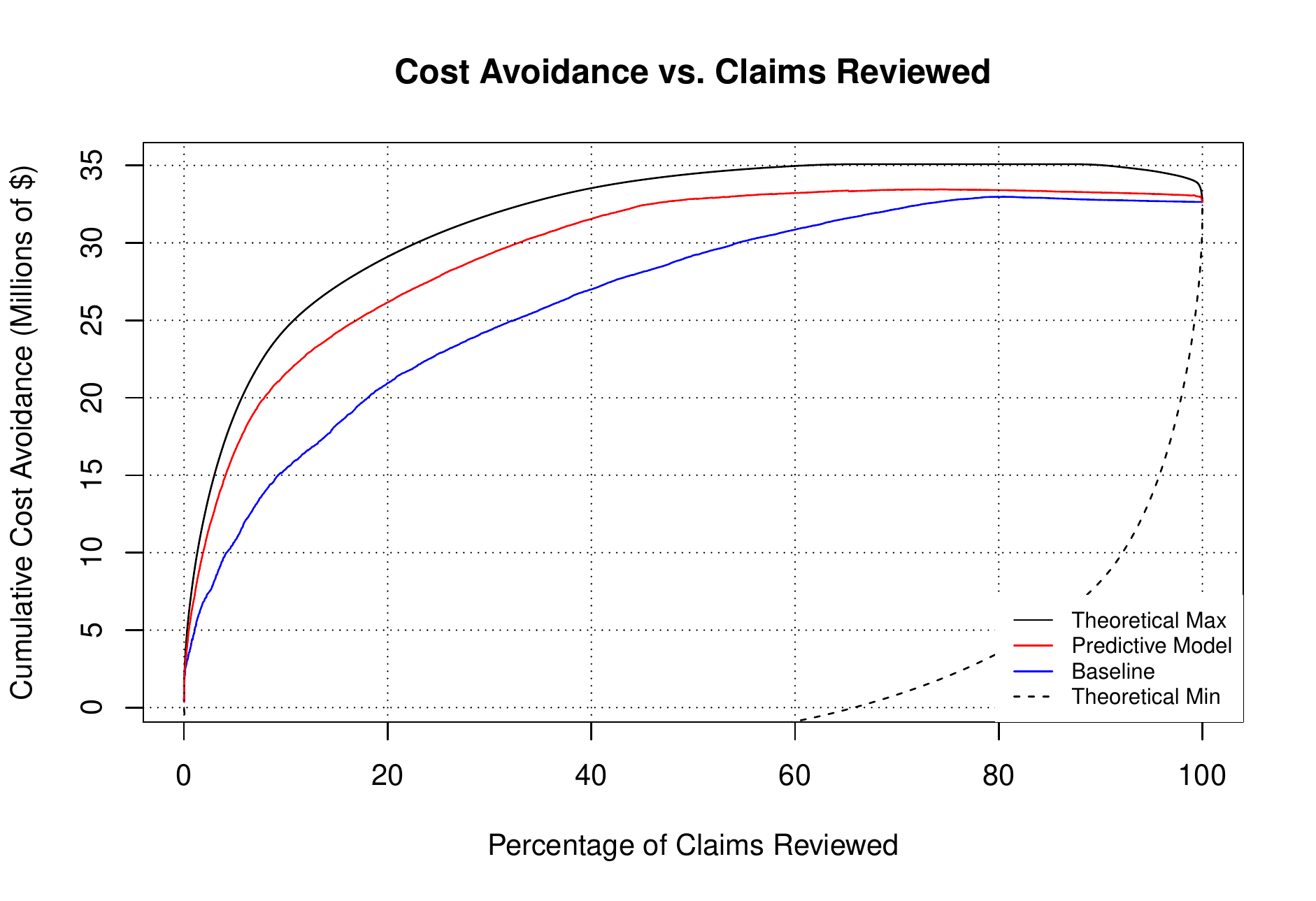}
\caption{\label{perfcurve}Cumulative cost avoidance as a function of percentage of claims reviewed under four different scenarios.}
\end{figure}

The red curve in Figure~\ref{perfcurve} displays the cumulative cost avoidance achieved by the predictive model vs.\ the percentage of claims reviewed.  Initially, as highly anomalous claims are reviewed, cost avoidance rises rather steeply, and then it levels off, eventually decreasing once negative value claims are encountered.

 \noindent  Three other scenarios are also presented to make comparisons.
\begin{enumerate}
\item
\textbf{Baseline.}  Claims are  sorted by billed amount.
\item
\textbf{Theoretical Maximum.}  Claims are sorted by a hypothetical model with perfect accuracy.
\item
\textbf{Theoretical Minimum.}   Claims are sorted by a hypothetical model that deliberately prioritizes low-value claims.
\end{enumerate}

\begin{table}
\begin{center}
{
\begin{tabular}{r|rr|r|c}
  \hline
  Reviewed (\%)  &  Baseline & Model &  \multicolumn{2}{c}{Improvement Over Baseline}  \\ 
  \hline
  10 & \$15.4M & \$21.6M &  \$6.2M & 40\% \\ 
   20 & 21.0 & 26.2 &  5.2 & 25\%   \\ 
   30 & 24.4 & 29.3 & 4.9  & 20\%  \\ 
   40 & 27.0 & 31.5 &  4.5 & 17\%  \\ 
   50 & 29.2 & 32.8 &  3.6 & 12\%  \\ 
   \hline
\end{tabular}}
\end{center}

\medskip

\caption{\label{perftable}Comparison of the cost avoidance model to baseline performance.}
\end{table}

\begin{figure}
\includegraphics[width=\textwidth]{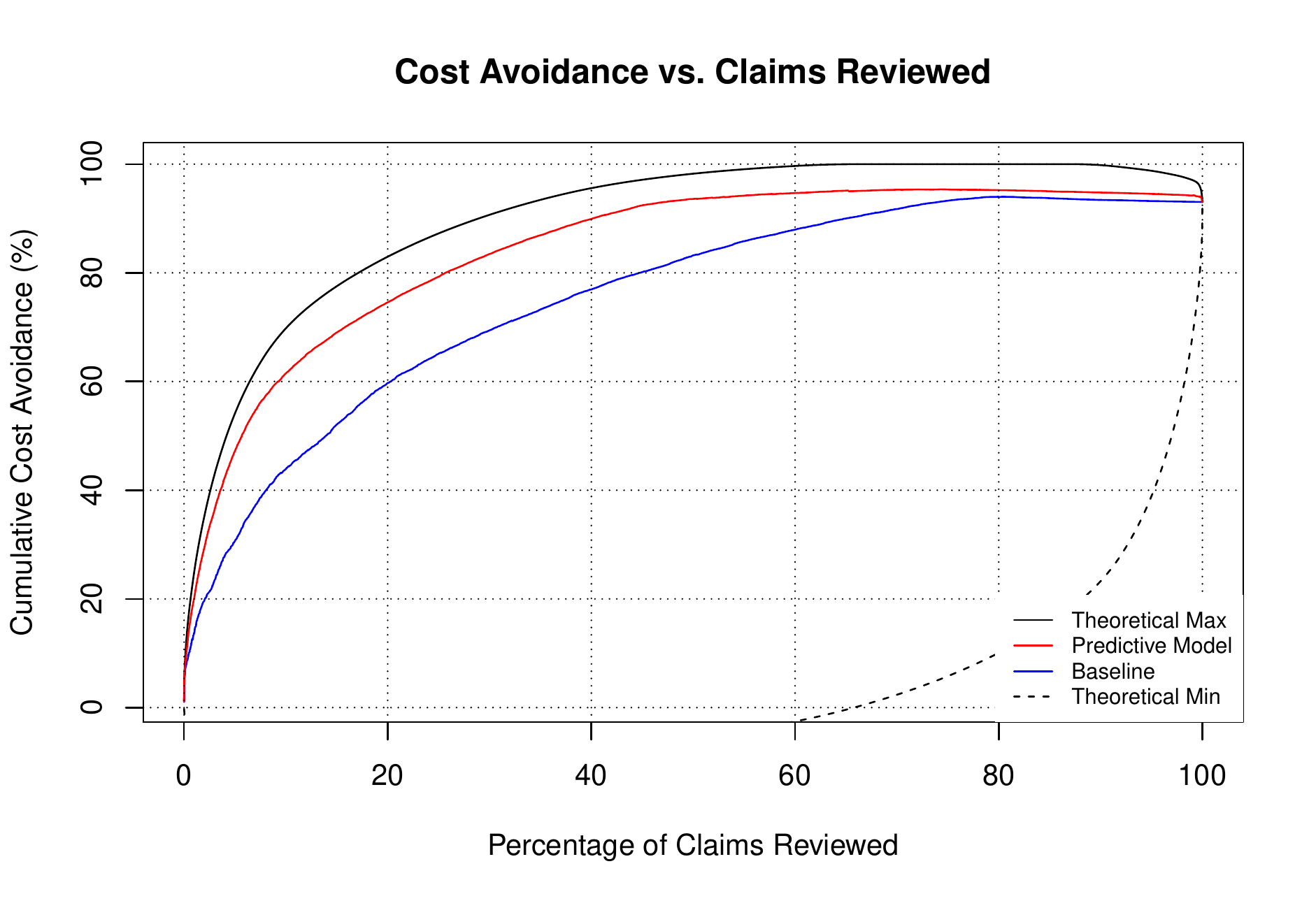}
\caption{\label{perfcurvepercent}Cumulative cost avoidance, expressed as a percentage of  potential savings, under four different scenarios. }
\end{figure}

\noindent  These curves reveal that the cost avoidance model approaches the theoretical maximum and significantly exceeds baseline performance.  Optimizing savings requires accounting for the cost of claims reviews, and  ideally most savings can be achieved by reviewing 50\% of claims or less.  In this range, the cost avoidance model saves 12\% to 40\% over baseline (see Table~\ref{perftable}).  Figure~\ref{perfcurvepercent} demonstrates that the model achieves 94\% of potential savings when 50\% of claims are reviewed, bordering the theoretical maximum of 98\% and substantially exceeding the baseline of 83\%.   These performance statistics are based on the previously mentioned sample of $n=\text{20,086}$ ER claims.  To put these savings into perspective, note that there were over  144 million ER visits in the United States in 2016, and approximately 132 million (92\%) of these visits were by insured patients \cite{Hcup, Sing2019}.  As a back-of-the-envelope approximation, if the  statistical distribution of relevant variables for national claims are approximately equal to those in this sample, adopting a cost avoidance optimization model nationally could save \$24B to \$41B over baseline annually.  This is a coarse approximation, but it suggests the potential savings available are likely in the tens of billions of dollars per year.

%
%

\subsection{Model Interpretation and Visualization}
\label{intvissect}  

\begin{figure}
\begin{center}
\includegraphics[width=12cm]{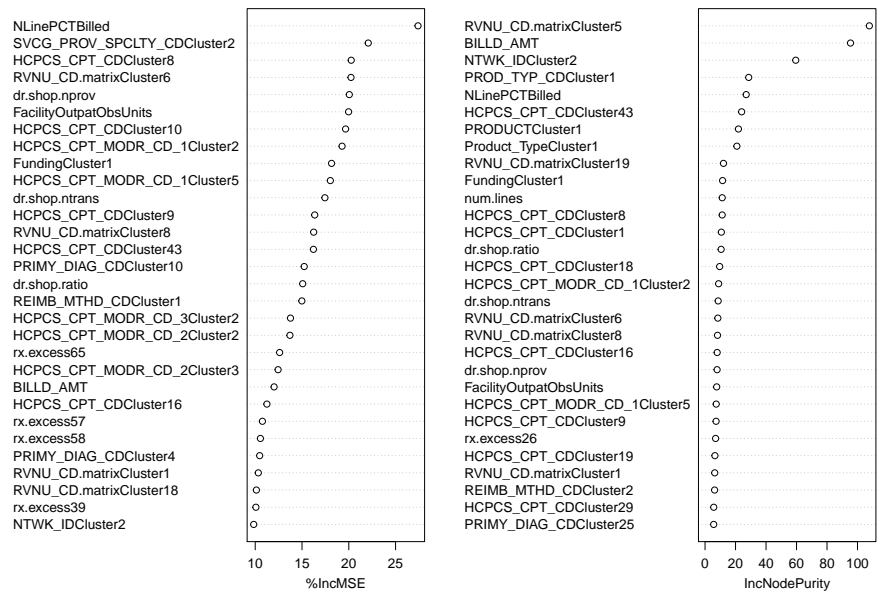}
\end{center}

\vspace{-7mm}

\caption{\label{impplot} Variable importance plots for the cost avoidance model.}

\vspace{5mm}
\end{figure}

Sections~\ref{rfsect} and \ref{econperfsect} demonstrated how to build a cost avoidance model and evaluate its overall performance.  We now turn our attention to ranking the relative importance of the predictor variables and visualizing them graphically.  Random forests achieve this objective with two variable importance metrics, increased mean square error (MSE) and increased node purity \cite{Breiman2001}.  The first of these measures the difference in MSE caused by permuting the data for each predictor $x_j$, normalized by the standard deviation of these differences.   The second metric is  the total decrease in node impurities from splitting on the variable, averaged over all trees in the random forest.  The variable importance plots for the cost avoidance model are displayed in Figure~\ref{impplot}.

\bigskip

\begin{figure}
\begin{center}
\includegraphics[width=12cm]{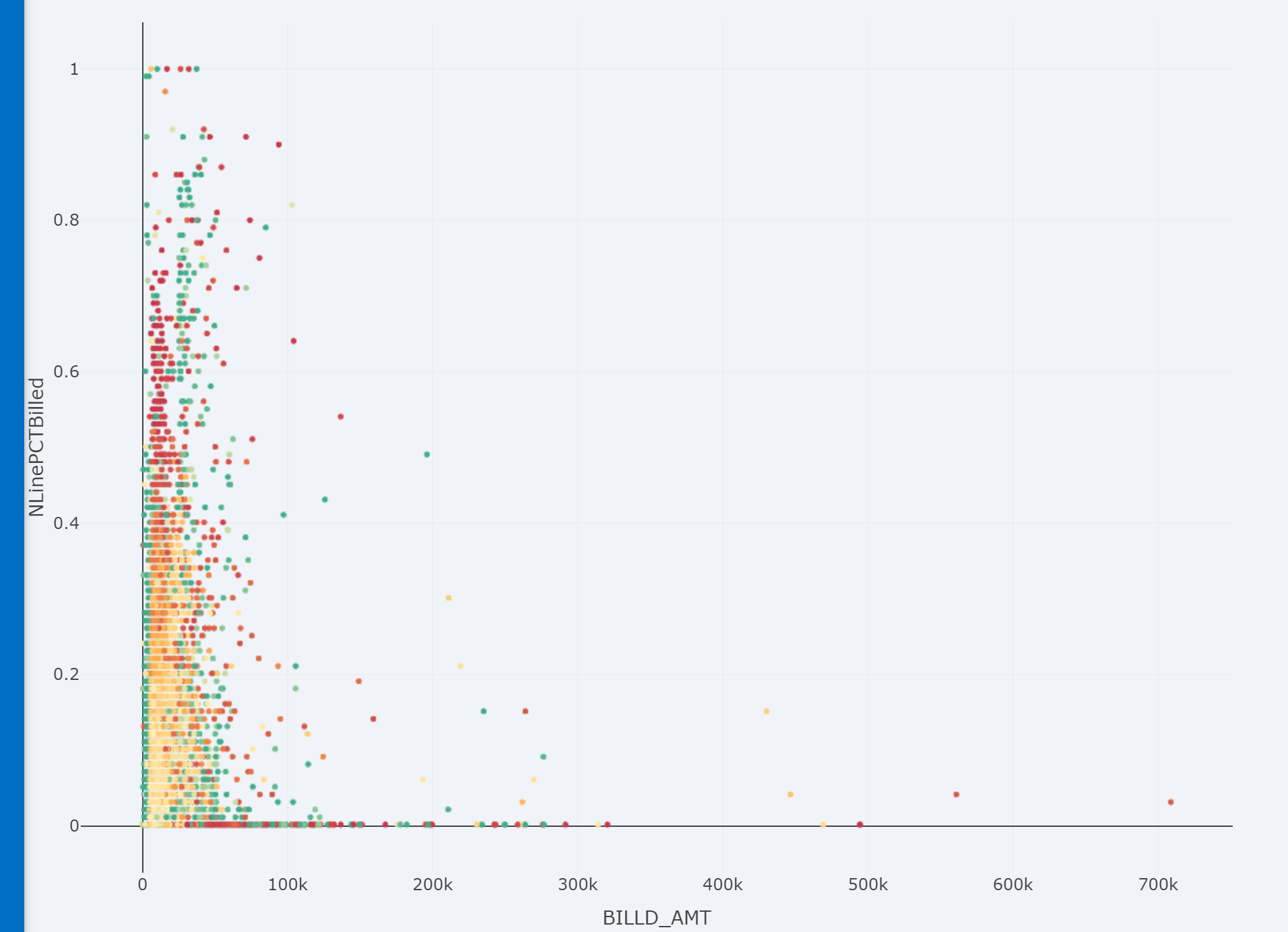}
\end{center}

\vspace{-2mm}

\caption{\label{heatmap} Anomaly score heat map, showing that claims with a higher percentage of financial carve-outs (N-lines) are more anomalous on average.}
\end{figure}

The most important variable, as measured by increased MSE, is the percentage of N-lines on a claim, referring to an institutional pricing code for transactions that are subject to financial carve-outs in a contract.  Unsurprisingly, claims with a disproportionately high percentage of N-lines are more likely to be anomalous, as confirmed by the heat-map in Figure~\ref{heatmap}.


The hierarchical clusters developed for categorical variables (e.g., HCPCS, Revenue, and Diagnosis Codes) appear prominently in the variable importance plots, as do churn metrics, such as facility observational units, unnecessary unbundling, and doctor shopping.  Another notable  predictor is prescription drug abuse, specifically for the following General Product Identifier (GPI) codes:

\begin{itemize}
\item
Opioids (GPI 65)
\item
Antianxiety Agents (GPI 57)
\item
Antidepressants (GPI 58)
\item
Antihyperlipidemics (GPI 39)
\item
Progestins (GPI 26) 
\end{itemize}


%
%
%

\section{Acknowledgments}

\noindent  We would like to thank Blue Cross and Blue Shield of Texas for funding this research under the Affordability Cures$^{\text{SM}}$ Initiative.

\bigskip

\bigskip

\bigskip

\bibliographystyle{unsrt}
\nocite{*}
\bibliography{ERClaimsAnomalyDetection}

\begin{thebibliography}{10}

\bibitem{GAO}
K.M. King.
\newblock Medicare fraud: Progress made, but more action needed to address
  medicare fraud, waste, and abuse, 2014.
\newblock \url{https://www.gao.gov/products/gao-14-560t}.

\bibitem{Che2013}
Ngufor Che and Wojtusiak Janusz.
\newblock Unsupervised labeling of data for supervised learning and its
  application to medical claims prediction.
\newblock {\em Computer Science}, 14(3):191, 2013.

\bibitem{Bauder2016}
Richard Bauder, Taghi~M. Khoshgoftaar, and Naeem Seliya.
\newblock A survey on the state of healthcare upcoding fraud analysis and
  detection.
\newblock {\em Health Services and Outcomes Research Methodology},
  17(1):31--55, July 2016.

\bibitem{He1997}
Hongxing He, Jincheng Wang, Warwick Graco, and Simon Hawkins.
\newblock Application of neural networks to detection of medical fraud.
\newblock {\em Expert Systems with Applications}, 13(4):329–336, 1997.

\bibitem{Rosenburg2000}
Marjorie~A. Rosenberg, Dennis~G. Fryback, and David~A. Katz.
\newblock A statistical model to detect {DRG} upcoding.
\newblock {\em Health Services and Outcomes Research Methodology},
  1(3/4):233–252, 2000.

\bibitem{Yamanishi2004}
Kenji Yamanishi, Jun-Ichi Takeuchi, Graham Williams, and Peter Milne.
\newblock On-line unsupervised outlier detection using finite mixtures with
  discounting learning algorithms.
\newblock {\em Data Mining and Knowledge Discovery}, 8(3):275–300, 2004.

\bibitem{Luo2010}
Wei Luo and Marcus Gallagher.
\newblock Unsupervised {DRG} upcoding detection in healthcare databases.
\newblock {\em 2010 IEEE International Conference on Data Mining Workshops},
  2010.

\bibitem{Thornton2013}
Dallas Thornton, Roland~M. Mueller, Paulus Schoutsen, and Jos {van
  Hillegersberg}.
\newblock Predicting healthcare fraud in medicaid: A multidimensional data
  model and analysis techniques for fraud detection.
\newblock {\em Procedia Technology}, 9:1252--1264, 2013.
\newblock CENTERIS 2013 - Conference on ENTERprise Information Systems /
  ProjMAN 2013 - International Conference on Project MANagement/ HCIST 2013 -
  International Conference on Health and Social Care Information Systems and
  Technologies.

\bibitem{Johnson2015}
Marina~Evrim Johnson and Nagen Nagarur.
\newblock Multi-stage methodology to detect health insurance claim fraud.
\newblock {\em Health Care Management Science}, 19(3):249–260, 2015.

\bibitem{cmsicd}
Medicare coverage general information {ICD}-10.
\newblock {\em Centers for Medicare and Medicaid Services}, 2021.
\newblock \url{https://www.cms.gov/Medicare/Coverage/CoverageGenInfo/ICD10}.

\bibitem{cmscpt}
{HCPCS} - general information.
\newblock {\em Centers for Medicare and Medicaid Services}, 2021.
\newblock \url{https://www.cms.gov/Medicare/Coding/MedHCPCSGenInfo}.

\bibitem{Bauder2017}
Richard Bauder, Taghi~M. Khoshgoftaar, and Naeem Seliya.
\newblock A survey on the state of healthcare upcoding fraud analysis and
  detection.
\newblock {\em Health Services and Outcomes Research Methodology}, 17:31--55,
  2017.

\bibitem{Sulc2015}
Zdenek Sulc and Hana Řezanková.
\newblock Dimensionality reduction of categorical data: Comparison of {HCA} and
  {CATPCA} approaches.
\newblock 2015.

\bibitem{Forgy1965}
Edward~W. Forgy.
\newblock Cluster analysis of multivariate data: efficiency versus
  interpretability of classifications.
\newblock {\em Biometrics}, 21(3):768–769, 1965.

\bibitem{Ester96adensity-based}
Martin Ester, Hans-Peter Kriegel, Jörg Sander, and Xiaowei Xu.
\newblock A density-based algorithm for discovering clusters in large spatial
  databases with noise.
\newblock pages 226--231. AAAI Press, 1996.

\bibitem{Tan2018}
Pang-Ning Tan, Michael Steinbach, Anuj Karpatne, and Vipin Kumar.
\newblock {\em Introduction to Data Mining, 2nd ed.}
\newblock Pearson, 2018.

\bibitem{Free2009}
David~A. Freedman.
\newblock {\em Statistical Models, Theory and Practice, Revised ed.}
\newblock Cambridge University Press, 2009.

\bibitem{OrdAUC}
Pelin Yıldırım, Ulaş~K. Birant, and Derya Birant.
\newblock {EBOC}: Ensemble-based ordinal classification in transportation.
\newblock {\em Journal of Advanced Transportation}, 2019.

\bibitem{Morgan1963}
James Morgan and John Sonquist.
\newblock Problems in the analysis of survey data, and a proposal.
\newblock {\em Journal of the American Statistical Association},
  58(302):415--434, 1963.

\bibitem{Breiman1984}
L.~Breiman, J.H. Friedman, R.A. Olshen, and C.J. Stone.
\newblock {\em Classification And Regression Trees}.
\newblock Routledge, 1984.

\bibitem{rfimg}
Afroz Chakure.
\newblock Random forest regression.
\newblock 2019.

\bibitem{Breiman2001}
Leo Breiman.
\newblock Random forests.
\newblock {\em Machine Learning}, 45(1), 2001.

\bibitem{Hcup}
Trends in emergency department visits.
\newblock {\em Healthcare Cost and Utilization Project. Agency for Healthcare
  Research and Quality}, 2021.
\newblock \url{https://www.hcup-us.ahrq.gov/faststats/NationalTrendsEDServlet}.

\bibitem{Sing2019}
Adam~J. Singer, Jr~Thode, Henry~C., and Jesse~M. Pines.
\newblock {US Emergency Department Visits and Hospital Discharges Among
  Uninsured Patients Before and After Implementation of the Affordable Care
  Act}.
\newblock {\em JAMA Network Open}, 2(4), 2019.

\bibitem{Cearnal2013}
Lee Cearnal.
\newblock Electronic medical records link to upcoding under fire.
\newblock {\em Annals of Emergency Medicine}, 61(4), 2013.

\bibitem{Sparrow2020}
Malcolm~K. Sparrow.
\newblock {\em License to steal: how fraud bleeds Americas health care system}.
\newblock Routledge, 2020.

\end{thebibliography}

\end{document}